\documentclass[10pt,letterpaper]{article}

\usepackage{cogsci}

\cogscifinalcopy % Uncomment this line for the final submission 
 \usepackage[table,xcdraw]{xcolor}
\usepackage{graphicx}
\usepackage{listings}
\usepackage{booktabs}
\usepackage{multirow}
\usepackage{amsmath,amssymb}
\usepackage{algorithm2e} %added by Jonathan
\usepackage{pslatex}
\usepackage{apacite}
\usepackage{soul}
\usepackage{subcaption}
\usepackage{comment} % added by Jonathan
\usepackage{csvsimple}
\usepackage{float} % Roger Levy added this and changed figure/table
\usepackage[round]{natbib}
\usepackage{hyperref} %Added by Andrea
\usepackage{todonotes}

\usepackage{algorithmic} %added by Andrea
\usepackage{amsmath} %added by Andrea
\usepackage{subcaption} %added by Andrea 
\usepackage{comment} %added by Andrea
\usepackage{wrapfig}

\RestyleAlgo{ruled}

\title{PACE: Procedural Abstractions for Communicating Efficiently}
\author{\large \bf
    Jonathan Thomas \quad Andrea Silvi \quad Devdatt Dubhashi \quad Moa Johansson \\ 
    Chalmers University of Technology and University of Gothenburg
}

\begin{document}

\maketitle

\renewcommand\thefootnote{} % Remove the footnote number
\footnotetext{Correspondence to: \texttt{silvi@chalmers.se}.}
\renewcommand\thefootnote{\arabic{footnote}} % Restore footnote numbering

\begin{abstract}
%\todo{change abstract from icml to cogsci}\textbf{todo: rewrite abstract from icml to cogsci}
A central but unresolved aspect of problem-solving in AI is the capability to introduce and use abstractions, something humans excel at. Work in cognitive science has demonstrated that humans tend towards higher levels of abstraction when engaged in collaborative task-oriented communication, enabling gradually shorter and more information-efficient utterances. Several computational methods have attempted to replicate this phenomenon, but all make unrealistic simplifying assumptions about how abstractions are introduced and learned. Our method, Procedural Abstractions for Communicating Efficiently (PACE), overcomes these limitations through a neuro-symbolic approach. On the symbolic side, we draw on work from library learning for proposing abstractions. We combine this with neural methods for communication and reinforcement learning, via a novel use of bandit algorithms for controlling the exploration and exploitation trade-off in introducing new abstractions. PACE exhibits similar tendencies to humans on a collaborative construction task from the cognitive science literature, where one agent (the architect) instructs the other (the builder) to reconstruct a scene of block-buildings. PACE results in the emergence of an efficient language as a by-product of collaborative communication. Beyond providing mechanistic insights into human communication, our work serves as a first step to providing conversational agents with the ability for human-like communicative abstractions.

\end{abstract}

\section{Introduction}

% It argues that neither Sapir nor Whorf endorsed a strong linguistic relativist notion of language but only a much weaker thesis: that language, occasionally, shapes and re-organizes cognition from A defense of a weak linguistic relativist thesis
%Procedural tasks are often found in cooking and programming and require the execution of a sequence of actions to achieve a desired goal.
Procedural tasks such as cooking and programming require executing a sequence of actions to achieve a desired goal. A natural approach to reduce their complexity and improve generalisation to new tasks is to introduce abstractions for common sequences of actions \citep{solway2014optimal}. For example, in cooking, techniques such as sautéing or kneading serve as foundational building blocks that simplify complex recipes. Similarly, abstractions emerge in repeated communication between human dyads when collaborating on shared tasks \citep{krauss1964changes, hawkins2020characterizing, McCarthyBuilding21}. Over time, as new abstractions are introduced into the shared language, communication becomes more concise. This can improve cooperation and allow to reach the goal more easily.

\begin{figure}[t]
    \centering
\includegraphics[width=\columnwidth]{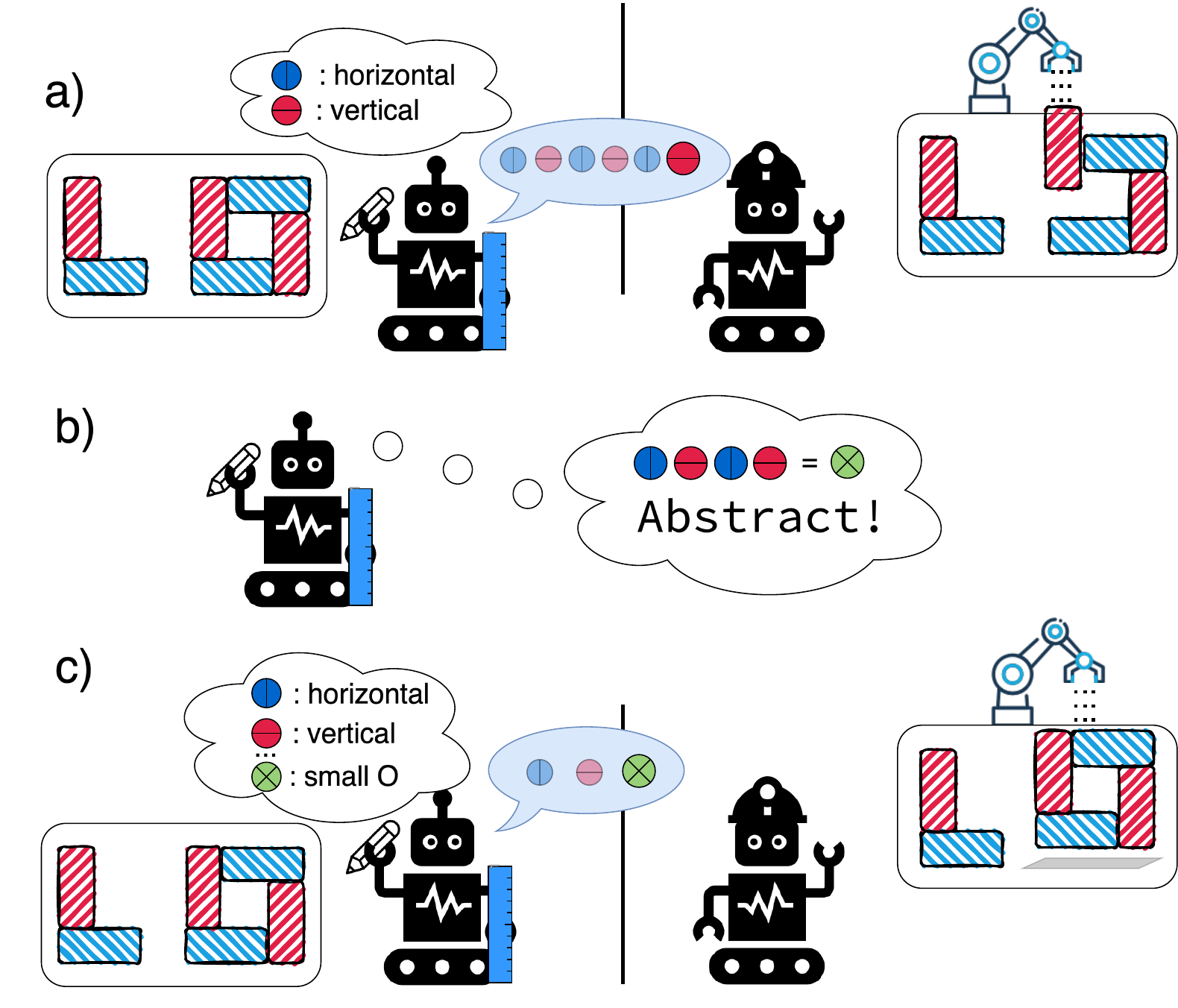}
%\includesvg[width=\columnwidth]{images/first_page_longer.svg}

    \caption{Two artificial agents playing the architect-builder game, starting from a small artificial language. Initially, the architect messages refer to horizontal or vertical blocks (a). After multiple interactions, the architect tries to introduce an abstraction (b), which after a learning period allows for shorter communication to solve the task (c).
    }
    \label{fig:arch_build_game}
\end{figure}

How communication shapes abstractions is of much interest within the AI and Cognitive Science communities \citep{lee1996whorf, HO2019}. A compelling principle is \textit{Efficient Communication}, which argues that languages are under pressure to be informative whilst minimising cognitive load \citep{Kemp2012, Gibson2016, Zaslavsky2019a, Gibson2019}. This provides insights into the abstractions humans converge to in semantic domains such as colour, kinships and others \citep{Xu2020, Regier2015, Kemp2012, yin24acl}. Here we want to explore the principle of Efficient Communication with artificial agents in collaborative procedural tasks: how do the pressures for efficient communication manifest and impact abstraction introduction and use?

%\cite{denic2024recursive} demonstrated how in other domains like recursive numeral systems, in which semantics are compositional and informativity can be maximal even with very few terms lexicalised, a different trade-off shapes languages: a pressure to lexicalise as few concepts as possible while simultaneously keeping utterances as brief as possible (i.e. minimize the average morphosyntactic complexity). Within collaborative procedural tasks, which pressures for efficient communication impact abstraction introduction and use and in what way? We take the first steps to answer this question within the context of artificial agents. 

%The interconnection between abstractions and their use in language is of much interest within Cognitive Science [sapir-whorf, lee]. In a paper by Ho et al. (2019) that combines perspectives from cognitive science, AI, and reinforcement learning, they suggest that “future work… needs to identify other pressures that can shape abstraction learning—such as the need to communicate and coordinate with others.” Building on this perspective, we investigate how communication impacts the formation and use of abstractions among conversational artificial agents working on cooperative procedural tasks.

% What is the architect-builder game
% Why other work sucks. 
% Emergent Communication, maybe? 

The architect-builder game introduced in \cite{McCarthyBuilding21} provides a simple framework to study the use of abstractions in collaborative tasks. It is a repeated game with two participants, the architect and the builder, who must complete a collaborative building task. In each round, the architect observes the goal-scene on their screen and conveys some instructions in natural language to the builder. The builder, who cannot see the original scene, interprets these instructions and attempts to reconstruct the shapes by placing blocks within the grid. The architect may send further instructions to help the builder. \citet{McCarthyBuilding21} showed that in early rounds human participants use lengthy utterances, while later realizing it is beneficial to introduce abstractions for commonly occurring shapes (e.g. an L-shape).  By doing this they move towards a more compact language, thereby leading to more efficient communication. Creating such abstractions is a key feature of human collaboration. In this work, we develop a computational framework of artificial agents to solve this kind of collaborative task (Figure \ref{fig:arch_build_game}). Previous work provides computational models of this process \citep{McCarthyBuilding21, jergeus2022towards}, but rely on simplifying assumptions, limiting the ability to capture the collaborative dynamics of language learning, which we address in this work.

We propose a novel multi-agent neuro-symbolic method called Procedural Abstractions for Communicating Efficiently (PACE). We integrate both neural and symbolic learning methods from the computer science literature. On the symbolic side, library learning \citep{ellis2021dreamcoder, bowers2023top}, a method from program synthesis that aims to abstract common subprograms into new, more easily reusable terms. On the neural side, emergent communication (EC) \citep{Foerster16, lazaridou2020emergent}, and reinforcement learning (RL) \citep{Sutton1998}, enable the development of a flexible, learnable communication language. PACE offers a unified framework for studying the formation and evolution of abstractions across multi-round interactions.

 In previous work, EC has been used to explain how communicative pressures for efficient communication shape the language structure in other settings, such as \citet{Carlsson21, carlsson2024cultural}. Here, we apply EC techniques to study how pressures manifest within abstraction learning. 

\begin{comment}
In this paper, we propose a novel multi-agent neuro-symbolic method called Procedural Abstractions for Communicating Efficiently (PACE). PACE integrates methods from library learning \cite{bowers2023top, ellis2021dreamcoder}, emergent communication (EC) \cite{lazaridou2020emergent, Foerster16}, and reinforcement learning (RL) \cite{sutton} and enables a unified approach to understanding abstraction formation and evolution within multi-round interactions. We apply PACE within an extended version of the Architect-Builder game. In our extension, the architect is a neuro-symbolic agent that composes programs to achieve the goal-scene within an internal symbolic language. The programs are neurally encoded as messages, following conventions from EC, and are then interpreted by the builder. After several rounds, the architect's internal symbolic language is extended with new abstractions via a probabilistic library learning method. These new abstractions provide alternative ways to express goal-scenes allowing for more concise programs and communication but require the agents to learn how to communicate these new abstraction. This contention between shorter but unestablished programs and more verbose established programs is handled via RL methods. 

\end{comment}
% Something about emergent communication. 
%Whilst EC has been used to study language \cite{lazaridou2020emergent}, 

We evaluate PACE on the Architect-Builder game, which we extend to artificial agents. In our extension, the architect composes programs to describe the goal-scene in an artificial symbolic language (Figure \ref{fig:arch_build_game} a). The programs are neurally encoded as messages, following conventions from EC, and are then interpreted by the builder, who tries to reconstruct the scene. After several rounds, the architect's internal symbolic language is extended with a new abstraction for a commonly occurring subprogram (Figure \ref{fig:arch_build_game} b). Each new abstraction provides alternative shorter ways to express goal-scenes, but require the agents to learn how to communicate and understand them. This contention between shorter but unestablished programs and more verbose established programs is handled via RL over repeated rounds of interaction (Figure \ref{fig:arch_build_game} c). We find that it exhibits similar tendencies to humans — a development toward a richer language that allows for more concise utterances. Interestingly, after a number of abstractions have been introduced, our model naturally converges to a stable language, after which no more abstractions are introduced. Moreover, we show that in this setting languages that are closer to optimality in terms of the trade-off between average morphosyntactic complexity and language size are easier to learn, connecting our work with the Efficient Communication literature. Our approach addresses limitations of existing approaches and provides a valuable framework for future exploration in this area.  

\section{Set-up: Architect-Builder Game}

In the Architect-Builder Game, the architect is provided with a set of goal-scenes depicting two adjacent shapes on a (9x9) grid. Each shape is a variable-size combination of 2x1 horizontal and 1x2 vertical blocks. The architect needs to communicate instructions to the builder (who does not know the goal state) that allow it to construct the goal-scene starting from a blank grid. 
This is inspired by the human experiments in \cite{McCarthyBuilding21}.
\paragraph{Dataset}
%(move the dataset description here under an explicit heading)
%We compose the goal-scene dataset similarly to \citet{McCarthyBuilding21}, where 
 We extend the dataset from \cite{McCarthyBuilding21} for our experiments. We increase the size going from $3$ to $31$ unique shapes, with multiple sub-shapes reoccurring in different shapes. Our shapes are of different sizes and resemble either uppercase or lowercase letters from the English alphabet. %(see Figure \ref{fig:dataset_shapes} in the Appendix). 
 As before, the dataset consists of scenes composed of two shapes placed side by side. Our dataset contains $961$ goal-scenes (compared to the original $9$), which enables us to use it for training neural agents. We split the dataset into 930 training scenes and 31 test scenes. These splits are constructed to ensure the distribution over shapes is the same. However, in the test set, the ordered pair of shapes constituting each goal-scene, does \emph{not} appear in the training set.
 
\paragraph{The agents}
The \emph{architect} is a neuro-symbolic agent: building instructions are constructed symbolically and encoded and communicated neurally. The \emph{builder} is a purely neural agent and its task is to learn to decode instructions to reconstruct the scenes step-by-step.
The architect has an internal symbolic action language ${\cal A}$ for constructing programs $p$ of building instructions for goal scenes. Initially, the language has only two primitives: ${\cal A}_{init} = \{a_{horiz}, a_{vert}\}$, corresponding to placing either a horizontal or vertical block. A \textit{program} of length $l$ takes the form $p=[a_1, ..., a_l]$, where initially, each $a_i$ is one of $\{a_{horiz}, a_{vert}\}$ \footnote{We note that along with each action $a_i$  we also attach positional information. As our goal is to learn abstractions of shapes, positions are irrelevant. Hence we assume positional encodings are pre-determined and omit them from further notation.}.

%To choose a program for a given scene, the architect is initialised with a pre-defined symbolic policy $\pi_{arch}$, which initially consists of a table of programs written in ${\cal A}_{init}$ for each scene in the training set together with a bandit...
%The task of the architect and builder is then to learn neural communication policies $\pi_{comm}$ and $\pi_{bldr}$, respectively. These policies are...

%The agents are trained together to learn policies that enable the architect to communicate instructions to the builder that enable it to reach the goal-state. We refer to the architect and builder's communication policies as $\pi_{comm}$ and $\pi_{bldr}$, respectively.

%\section{PACE: Procedural Abstractions for Communicating Efficiently}
\section{PACE: Selection, Communication and Abstraction}
\begin{figure}[t]
    \centering
\includegraphics[width=0.49\textwidth]{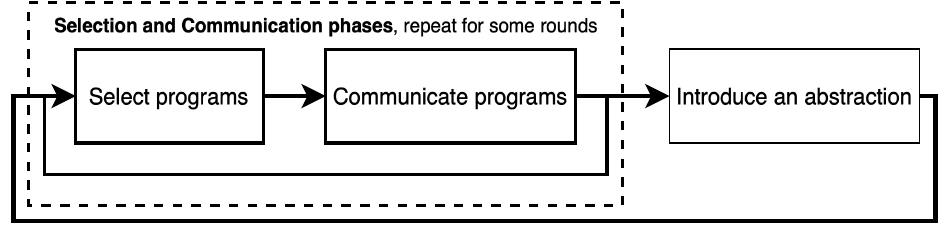}
%\includesvg[width=0.48\textwidth]{images/simple_algorithm_final.svg}

    \caption{Interaction between the architect and builder in PACE proceeds as follows: (1) given a goal-scene the architect chooses the program to communicate, and (2) the architect and builder communicate via EC. After multiple interactions, an abstraction is introduced. Then the loop repeats.
    }
    \label{fig:pace_architecture}
\end{figure}

There are three phases to PACE, \textbf{Selection}, \textbf{Communication} and \textbf{Abstraction} (see Figure \ref{fig:pace_architecture}).
 The architect must first choose a program to communicate. For this purpose the architect is initialised with a table of programs written in ${\cal A}_{init}$ for each scene in the training set. Initially there is one program per scene. Secondly, the architect learns how to communicate (neurally) instructions to the builder via EC, allowing them to learn a language for collaboration.
After some rounds, the architect enters the Abstraction phase, where it uses a library learning mechanism to identify common sub-sequences to abstract. The new abstraction is then used to add additional (shorter) programs for the relevant scenes. 
In the next Communication phase, the architect thus has multiple programs to choose from. We use reinforcement learning techniques to learn to choose between alternative programs. We next describe these three phases in more detail. We also provide more detailed algorithms of PACE in the Appendix.

%At a high-level interaction between the architect and builder in PACE proceeds as follows: (1) given a goal-scene the architect chooses a program via the bandit, and (2) the architect and builder communicate via EC. After multiple interactions, an abstraction is introduced through library learning. Then the loop repeats. Appendix \ref{app:algos} presents more detailed algorithms explaining each phase.

%identifies new abstractions, these are common sub-sequences within programs that are introduced through a library learning mechanism. In Communication component involves neural communication between the architect and builder using techniques from EC to learn a language for collaboration. The Exploration/Exploitation component utilises bandit techniques to choose between alternative programs for a given goal-scene.

\begin{comment}
    
There are three main components to ACE:
\begin{description}
    \item[Abstraction] The architect identifies new abstractions: these are common sub-sequences within programs that are introduced through a library learning mechanism. 
    \item[Communication] The (neural) communication between the architect and builder, using techniques from EC to learn a language for collaboration.
    \item[Exploration/Exploitation] Bandit techniques to choose between alternative programs for a given goal-scene.
\end{description}

\end{comment}

\paragraph{Selecting programs}
The architect initially has one program to construct each scene from an empty grid.
As abstractions are introduced, there will be multiple alternative programs of different lengths available to the architect. They may differ in reconstruction accuracy, in particular, programs containing a new abstraction will initially have lower accuracy, as the builder has not been exposed to them. Hence, this is a classic \textit{exploration} vs. \textit{exploitation} problem for which reinforcement learning with \emph{bandit techniques}\footnote{an analogy coming from a gambler repeatedly having to select which arm of a slot machine to play.} are well-suited \citep{Sutton1998, lattimore2020bandit}
%contextual = each goal scene has different arms.
%Arm = alternative program for a given scene.
% Q-values = quality of program calculated via quality of its individial actions (combinatorial).
% Reward = reconstruction accuracy
%Multi-armed bandit algorithms balance reward maximization by exploiting knowledge from previous trials while exploring new actions to discover potentially better knowledge.
More specifically, we view the selection between programs as a contextual multi-armed bandit with combinatorial actions: in general each goal-scene has multiple programs to choose from, corresponding to the arms of the bandit. To estimate the quality of each program, we maintain a table of Q-values, one for each action in $\mathcal{A}$, with the quality of the program being the product of the Q-values of its actions: $Q(p) = \prod_{i = 1}^{|p|} \gamma Q(a_i)$.\footnote{We empirically found that initialising the Q-values to $0$ works best in practice.} This captures the trade-off between program length and the communicative accuracy of instructions. We empirically determine a value of $\gamma=0.99$, which results in a small bias for shorter programs. To ensure that the architect also explores new programs, we adopt an $\epsilon$-greedy strategy, where with probability $1-\epsilon$ a random program (arm) is selected. We fix  $\epsilon$ at a constant value of $0.1$, ensuring a constant level of exploration.  %where $\epsilon$ is held at a constant value, ensuring a constant level of exploration.
After the communication of the program (see below) we update the estimated value of its component actions as $Q(a) \leftarrow Q(a) + \alpha ( r - Q(a))$,
where $r$ is the reconstruction accuracy which is 1 if this instruction was successfully interpreted by the builder and 0 otherwise.
This approach ensures that the new programs introduced after each Abstraction phase will be explored by the architect.

%We view the selection between programs as a contextual bandit with combinatorial actions \citep{lattimore2020bandit}, where the expected value of the program is obtained by combining the expected value of the actions in it.
%We update the estimated value of a single action as $Q(a) \leftarrow Q(a) + \alpha ( r - Q(a))$,
%where $r$ is the reconstruction accuracy which is 1 if the instruction is successfully interpreted by the builder and 0 otherwise. The parameter $\alpha$ is the learning rate.
%The value of an entire program is calculated according as $Q(p) = \prod_{i = 1}^{|p|} \gamma Q(a_i)$. This captures the trade-off between program length and the communicative accuracy of instructions. While we consider the value $\gamma$ in the interval $(0, 1]$, we end up using a value of $\gamma=0.99$, which results in a small bias for shorter programs. \textit{Exploration vs. exploitation} is undertaken using an $\epsilon$-greedy strategy, where $\epsilon$ is held at a constant value, ensuring a constant level of exploration.
%Through this, we allow the architect to control the \emph{regret}, i.e. the difference between the optimal and observed reward, arising from the introduction of new abstractions. 

\paragraph{Communicating a program}
\label{subsec:comms}

%\textcolor{black}{Human languages are subject to pressures to minimise cognitive load and be either informative or minimise the utterances' morphosyntactic complexity in the case of compositional semantics. In previous work, EC has been used to explain the language structure in the numerals \citep{Carlsson21} and colours domains \citep{carlsson2024cultural}. We apply EC techniques to study how pressures manifest within abstraction learning.} 

Having selected a program, $p$, the architect and builder play a one-step signalling game for each of its instructions $a_i$. This takes the form of $(x_i,a_i,x_{i+1})$: the grid state $x_i$ is transformed into $x_{i+1}$ by action $a_i$. The architect learns a neural communication policy $\pi_{comm}$, which produces a message $m_i = \pi_{comm}(a_i)$. Similarly, the builder learns a policy $\pi_{bldr}$ which estimates the next grid-state, and is defined as $\widehat{x}_{i+1} = \pi_{bldr}(x_i, m_i)$. These are both implemented as fully-connected neural networks. Since our messages are discrete, we use the gumbel-softmax relaxation to sample from discrete messages which makes our model end-to-end differentiable \citep{jang2017categoricalreparameterizationgumbelsoftmax}. The policies are jointly trained to minimise the binary cross-entropy loss between the target next state $x_{i+1}$ and the builder's output $\widehat{x}_{i+1}$. We also introduce a bias for positive signalling \cite{eccles2019biases} in the architect's loss to encourage the architect's messages to carry meaningful information about the instruction they represent.

\paragraph{Introducing Abstractions}
After some rounds of communication, the architect enters the abstraction phase, where it searches for novel abstractions allowing for shorter programs for describing goal scenes.
% As the interaction proceeds with the builder, the architect may discover common abstractions which can shorten the communication. These can be used to rewrite current programs allowing for more compressed versions which are then available as an alternative in later interactions.
%The architect can analyse the available programs and introduce into the symbolic language ${\cal A}$ a new abstraction for a sub-sequence that enables the maximal reduction in the length of the whole set of programs. 
Abstractions are constructed by an improved version of the procedure used in \citet{McCarthyBuilding21}, where we are able to remove the explicit upper-size limit on the library size.
%which itself is 
%inspired by library learning methods like DreamCoder \citep{ellis2021dreamcoder}. Note that from \cite{McCarthyBuilding21} approach, we are able to remove the explicit constraint on the library size, which appears in their likelihood function.

The architect evaluates the set of candidate abstractions extracted from their programs %\footnote{We also allow for not modifying the language $\cal A$, if this is already maximal.}  
 and picks the one which maximises (\ref{eq:bayes_update}): 
\begin{equation} 
\label{eq:bayes_update}
    P(\mathcal{A} \cup \{a_{cand}\} | \{p_{i}\}_{i=1}^{N}) \propto  
    %P(\mathcal{A} \cup \{a_{cand}\}) \cdot 
    \prod_{i=1}^{N} P(p_{i}| \mathcal{A} \cup \{a_{cand}\})
\end{equation}
where $a_{cand}$ is the new candidate abstraction, $\{p_{i}\}_{i=1}^{N}$ are the known programs so far. The right side of the equation is further defined in \ref{eq:likelihood}. It provides a measure of the expected reduction in program length, which is defined in terms of the minimum description length ($\mathit{MDL}$) --the shortest program achievable using also the new candidate abstraction.

%This captures an inductive bias of the architect to introduce abstractions that are informative, as the probability of a program under the augmented language is proportional to the reduction in program length when re-writing the program using the candidate abstraction. 

\begin{equation}
\label{eq:likelihood}
    P(p_{i} | \mathcal{A} \cup \{a_{cand}\}) = exp(-\mathit{MDL}(p_{i}| \mathcal{A} \cup \{a_{cand}\}))
\end{equation}

\begin{figure*}[htbp]
  \centering
  \begin{subfigure}[b]{0.49\textwidth}
    \centering
    \includegraphics[width=\textwidth]{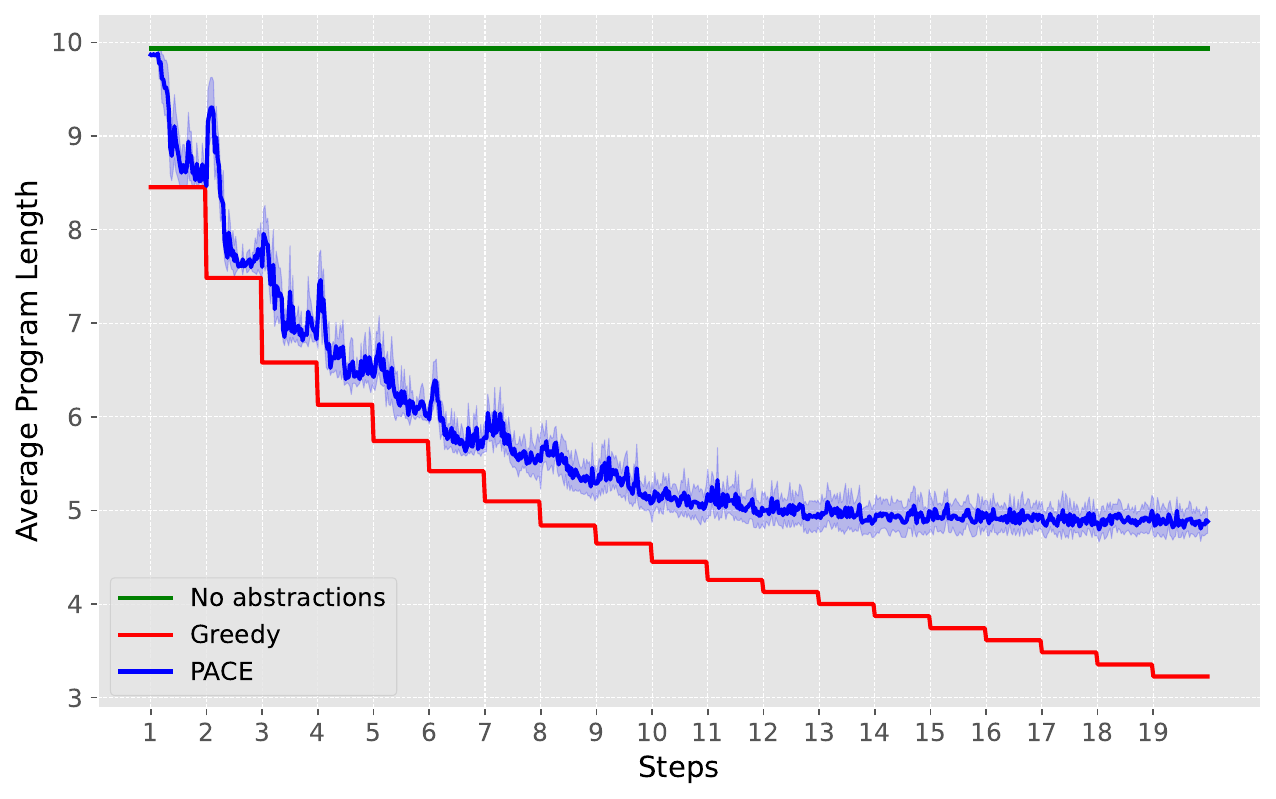}
  \end{subfigure}
  \hfill
  \begin{subfigure}[b]{0.49\textwidth}
    \centering
    \includegraphics[width=\textwidth]{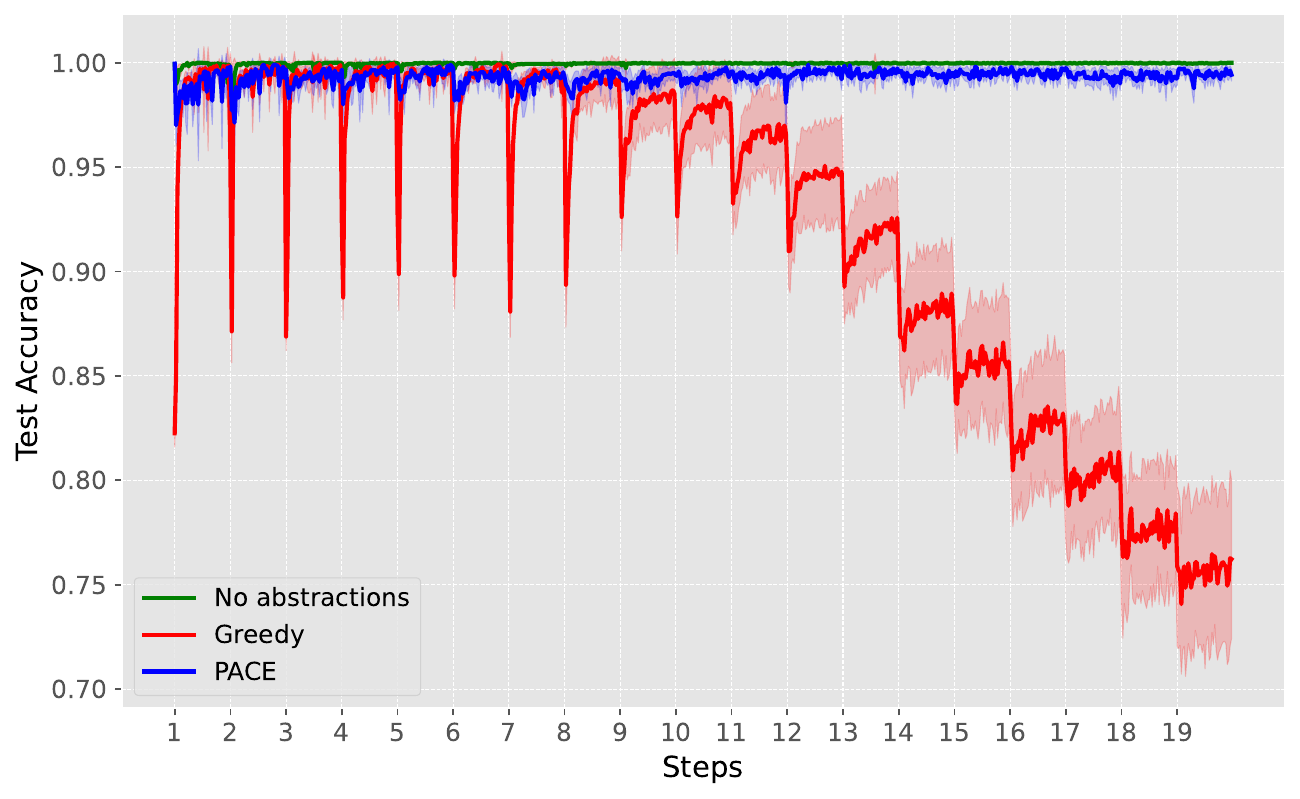}
    %, naive emergent communication that introduces abstractions (greedy) and an naive emergent communication that does not introduce abstractions (no abstractions). Line indicates mean performance and shaded regions indicate the 95\% confidence interval for $20$ random seeds.}
  \end{subfigure}
  \caption{Comparison between PACE, \emph{Greedy} and \emph{No abstractions} in terms of program length and test reward over time. Line indicates mean value and shaded regions indicate the 95\% confidence interval.}
  \label{fig:result}
\end{figure*}

\subsection{Efficient Communication in PACE}

As in other settings like recursive numeral systems \citep{denic2024recursive}, programs in the Architect-Builder game can be perfectly informative even with a minimal vocabulary -- precisely of size two, with one term referring to the horizontal primitive block and the other referring to the vertical one. With just the two initial action-words the architect can describe infinitely complex goal-scenes, given enough time! This is because the semantics of the environment are compositional. Thus, as \cite{denic2024recursive} show for recursive numeral systems, we expect that the pressures in competition to shape the agents' languages in the Architect-Builder game are the average morphosyntactic complexity of the programs describing the goal-scenes and the language size. In our case this corresponds to a trade-off between the average length of the programs communicated by the architect, and the number of lexicalised terms, i.e. the number of primitives plus abstractions in $\mathcal{A}$, the architect's lexicon.

%\makeatletter
%\renewcommand{\ALG@name}{Alg.}
%\makeatother

\subsection{Implementation Details}
\label{app:implement}

 The neural agents are deep neural networks which have $1$ and $2$ hidden layers of size $200$, respectively.  The architect's output layer is of size $30$, its maximum vocabulary size.  The output layer of the builder is of size $81$, corresponding to the size of the grid (9x9). We use the gumbel-softmax relaxation with the \emph{hard} parameter so that the networks can be differentiated end-to-end. We use a learning rate of $0.0009$ in conjunction with the ADAM optimiser \citep{kingma2014adam}. The bandit hyperparameters $\alpha$, $\gamma$, $\epsilon$ and $q_{init}$ are set to $0.5$, $0.99$, $0.1$, and $0.0$ respectively. Note, that before an abstraction is introduced we prune all but the $3$ best programs (determined empirically) for each goal-scene in the architects symbolic table to not incur an exponential growth in the number of programs for each goal-scene. We also set the number of epochs $e$ to 40.
All hyperparameters chosen are experimentally determined through grid search.

\section{Results \& Discussion}
\label{sec:eval}

Through empirical experimentation we want to investigate the following questions:
1) Does PACE display conversational tendencies which are similar to humans, allowing for more concise communication after multiple interactions? %(merge in 4.3 + 4.4 + 4.5)
2) Does communicative pressures impact what abstractions get adopted? If so, which abstractions are adopted and which are discarded? %4.2 but this is more about the size of the language. I'd say also Fig 2b is related to this.
%We conduct our experimentation in an extension of the Architect-Builder game -- a domain in which we expect communicative pressures to exist between average morphosyntactic complexity and vocabulary size. In this setting, humans converge towards specialised languages which reduce average morphosyntactic complexity through introducing more abstract terms. 
The behaviour of PACE is compared to two naïve baselines, to assess the impact of each of PACE components better. The first, \emph{No abstractions}, is PACE without the abstraction phase. The second, \emph{Greedy}, always picks the shortest program, eliminating the bandit. We use reward (also referred to as reconstruction accuracy) %cumulative regret, defined as $Regret_T = \sum_{t=1}^T \left( r^* - r_{t} \right)$ 
and average morphosyntactic complexity to compare these approaches. Unless otherwise stated results obtained are averaged over $16$ runs. %We provide further implementation details in the Appendix.

\paragraph{PACE reduces average morphosyntactic complexity}%Less regret, with more informativity}

PACE does indeed reduce morphosyntactic complexity as seen in Figure \ref{fig:result} (left). Starting from an average morphosyntactic complexity of $9.95$, over successive interaction steps, PACE is able to reduce average morphosyntactic complexity consistently and finally converges around a value of $4.92 \pm 0.20$ . 
%PACE does indeed converge to shorter programs over interaction steps, as seen in Figure \ref{fig:result} (left). 
When we compare PACE to the \emph{No abstraction} variant, the average morphosyntactic complexity is halved, going from an average of $10$, indicating that a more efficient language has been derived. The \emph{Greedy} variant reduces program lengths even more, however, this comes at a price. In Figure \ref{fig:result} (right) we see drastic drops in accuracy for \emph{Greedy} every time an abstraction is introduced, whereas PACE's slower introduction leads to smaller drops in accuracy. For a while, both strategies manage to recover near full accuracy, but after some time, the \emph{Greedy} communication fails to recover and becomes progressively worse.

\begin{figure}[t]
    \centering
\includegraphics[width=0.49\textwidth]{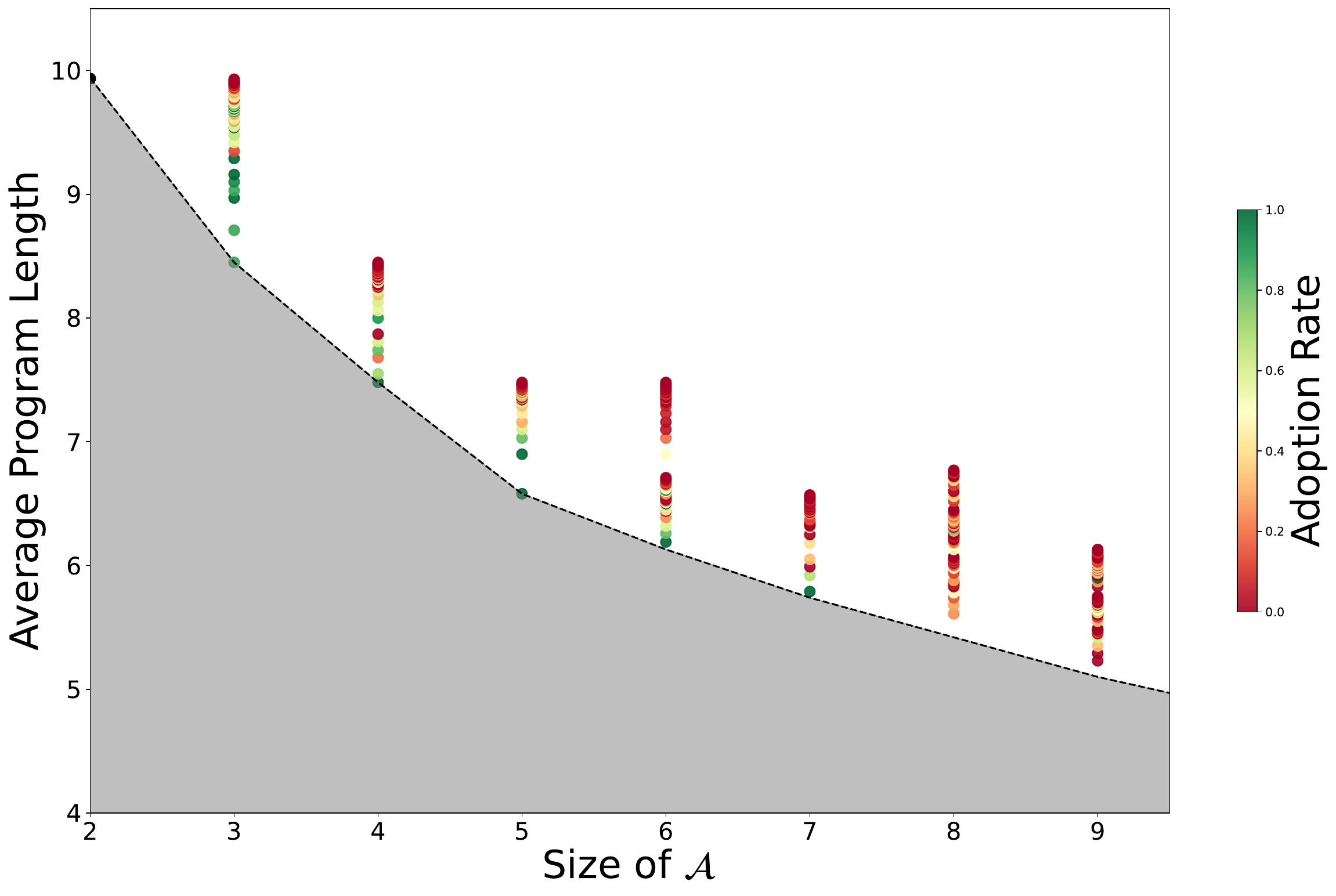}

    \caption{Adoption rate of possible abstractions by language size. As the language grows, fewer new abstractions are adopted. The dashed line represents the interpolation of the (discrete) Pareto Frontier calculated as the trade-off between average program length and size of the language $\mathcal{A}$. The grey area represents unachievable languages.
    %Comparison of different abstractions being introduced in $\mathcal{A}$ at different abstraction steps. We abstract shapes of different sizes and frequencies and, after some rounds of communication, we check if the abstraction is adopted or discarded by the architect. Each point represents the union of the language $\mathcal{A}$ and a new abstraction, plotted in terms of the vocabulary size and the resulting average program length, and the colour represents the probability of this abstraction being adopted. The dashed line represents the interpolation of the (discrete) Pareto Frontier calculated in terms of the trade-off between average program length and size of the language $\mathcal{A}$, while the grey area represents unachievable languages.
    }
    \label{fig:pareto_frontier}
\end{figure}
\begin{comment}
    
\begin{figure*}[t]
    \centering
%\includegraphics[width=\columnwidth]{images/first_page_library.pdf}
\includegraphics[width=\textwidth]{images/morpho_vs_size.pdf}

    \caption{IDK
    }
    \label{fig:pareto_frontier}
\end{figure*}
\end{comment}

\begin{figure*}[htb]
  \centering
  \begin{subfigure}[b]{0.49\textwidth}
    \centering
    \includegraphics[width=\textwidth]{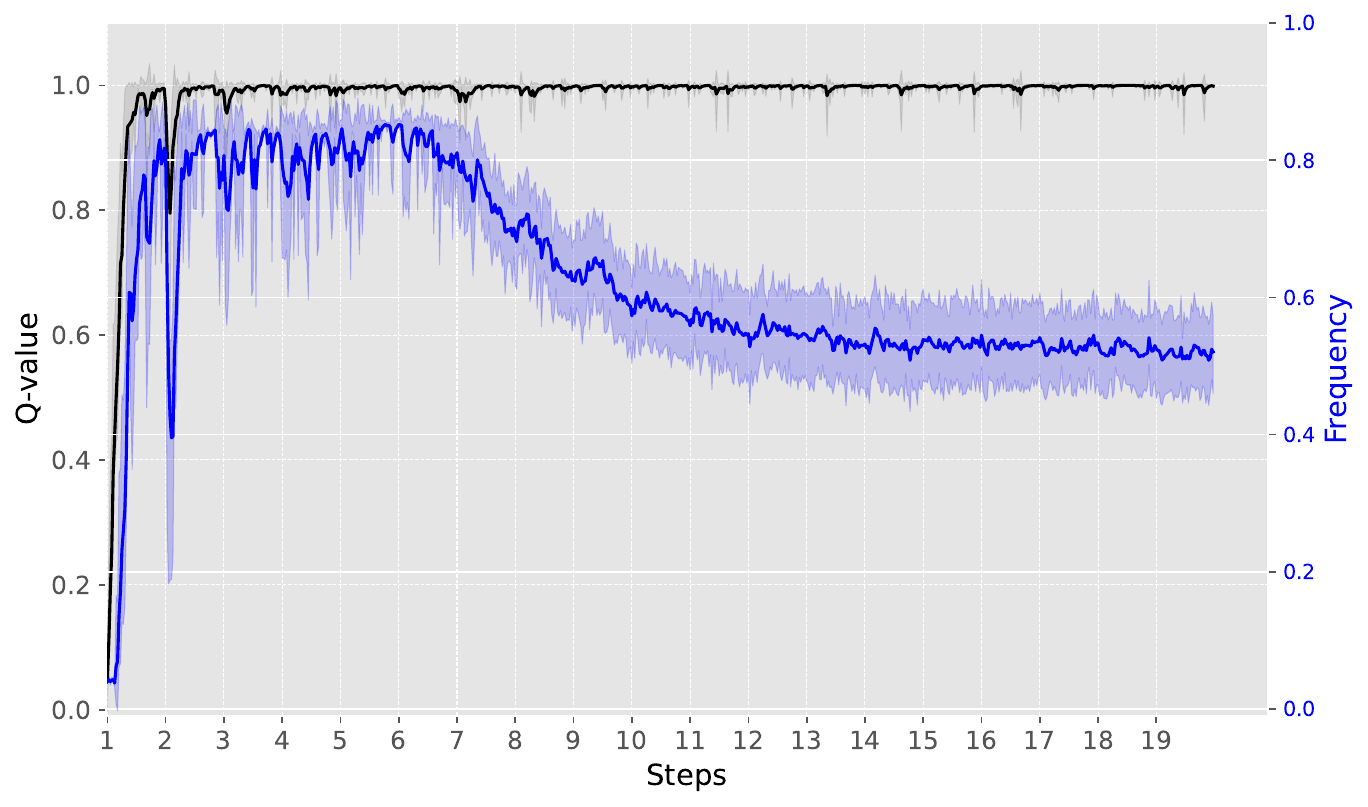}
    \caption{The tower abstraction being introduced successfully.}
    \label{fig:good_abstraction}
  \end{subfigure}
  \hfill
  \begin{subfigure}[b]{0.49\textwidth}
    \centering
    \includegraphics[width=\textwidth]{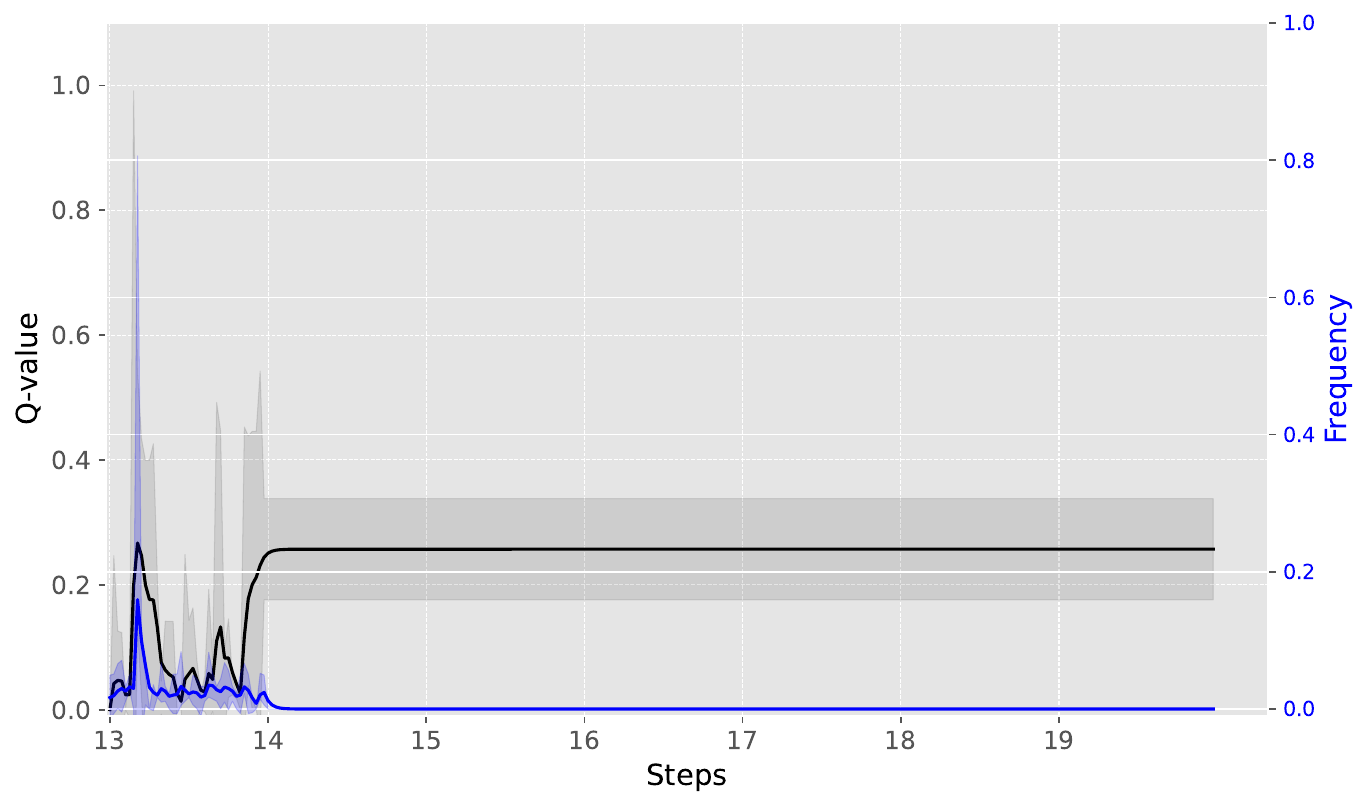}
    \caption{The rotated H abstraction gets eventually discarded.}
    \label{fig:bad_abstraction}
  \end{subfigure}
  \caption{Different abstractions being introduced. We show the mean and 95\% confidence interval for Q-value (black) and frequency (blue) versus epochs (of which there are $40$ in a step).}
  \label{fig:introducing_abstractions}
\end{figure*}

\paragraph{Why does PACE converge?}

As more terms are introduced and the average morphosyntactic complexity reduces, we find that it becomes harder to introduce new abstract terms. This explains why the \emph{Greedy} approach diverges as it is forced to use new abstractions in its programs. This raises a profound question which is of interest within efficient communication: why does PACE introduce some abstractions whereas others are omitted? To understand this we conduct an experiment where at each step, before the communicative phase, we gather all candidate abstractions, group them for their size (as in how big the shape the abstraction refers to is in terms of primitive blocks) and frequency in our dataset, and then introduce them one by one into $\mathcal{A}$. For each candidate abstraction, we let the architect and builder communicate until convergence and then probe the final language to check if the tested abstraction has been adopted or not. \footnote{Note that to make this experiment computationally tractable we limit to sampling $3$ abstractions for each frequency and size, wherever we have more than $3$.} This provides an exhaustive search of differing abstractions and allows us to analyse their ease of adoption or \emph{learnability} into the current language \citep{steinert2020ease}, based on how much reduction in average morphosyntactic complexity they can potentially lead to. At the next step, the experiment is repeated with the new symbolic language $\mathcal{A}$ augmented with the new abstraction introduced at the previous Abstraction phase, making each step of this experiment conditioned on the current language of PACE.

We plot the results in Figure \ref{fig:pareto_frontier}. We find that when $\mathcal{A}$ is only composed of primitive terms (the language is small), abstractions are generally always adopted but as $\mathcal{A}$ grows abstractions which would result in languages that are further away from the Pareto Frontier become harder to learn. Eventually, introducing new abstractions becomes infeasible, resulting in PACE's convergence. This demonstrates how communicative efficiency impact language formation within PACE, and connects with \citet{steinert2020ease} which show that simplicity and ease of learning are intertwined.  

\begin{comment}
    
\begin{figure}[h]
    \centering
    \includegraphics[width=\textwidth]{images/reconstruct_model/cumulative_regret_vs_steps.pdf}
    \caption{Cumulative regret vs steps.}
    \label{fig:cumulative_regret}
\end{figure}

\end{comment}

\paragraph{Which Abstractions are Adopted?}

\begin{comment}
    
\begin{figure}[htbp]
\centering
\includegraphics[width=\linewidth]{images/reconstruct_model/abstraction_1_introduction.pdf}
\caption{The first abstraction, a tower, being introduced successfully.}
\label{fig:good_abstraction}
\end{figure}

\begin{figure}[htbp]
\centering
\includegraphics[width=\linewidth]{images/reconstruct_model/abstraction_13_introduction.pdf}
\caption{The fourteenth abstraction, resembling an X, is introduced, but not retained.}
\label{fig:bad_abstraction}
\end{figure}

\end{comment}

As discussed, PACE's final language does not include all abstractions suggested. In Figure \ref{fig:introducing_abstractions}, we show some that experience different fates. Figure \ref{fig:introducing_abstractions} (left) shows the Q-value and frequency for the tower abstraction, consisting of two stacked vertical blocks, which is retained in PACE's final language. Figure \ref{fig:introducing_abstractions} (right) instead shows the same for a more complex abstraction resembling an H rotated by 90 degrees, which after a trial period was not retained, judged as too hard to be learnt by the builder and thus discarded by the architect via the bandit. This is connected to the different frequencies of these two shapes in our dataset, with the tower being more than 20 times more common than the rotated H, resulting in the first one being easier for the builder to understand and ultimately adopted by the architect into their language.

\begin{figure}[h]
    \centering
    \includegraphics[width=\linewidth]{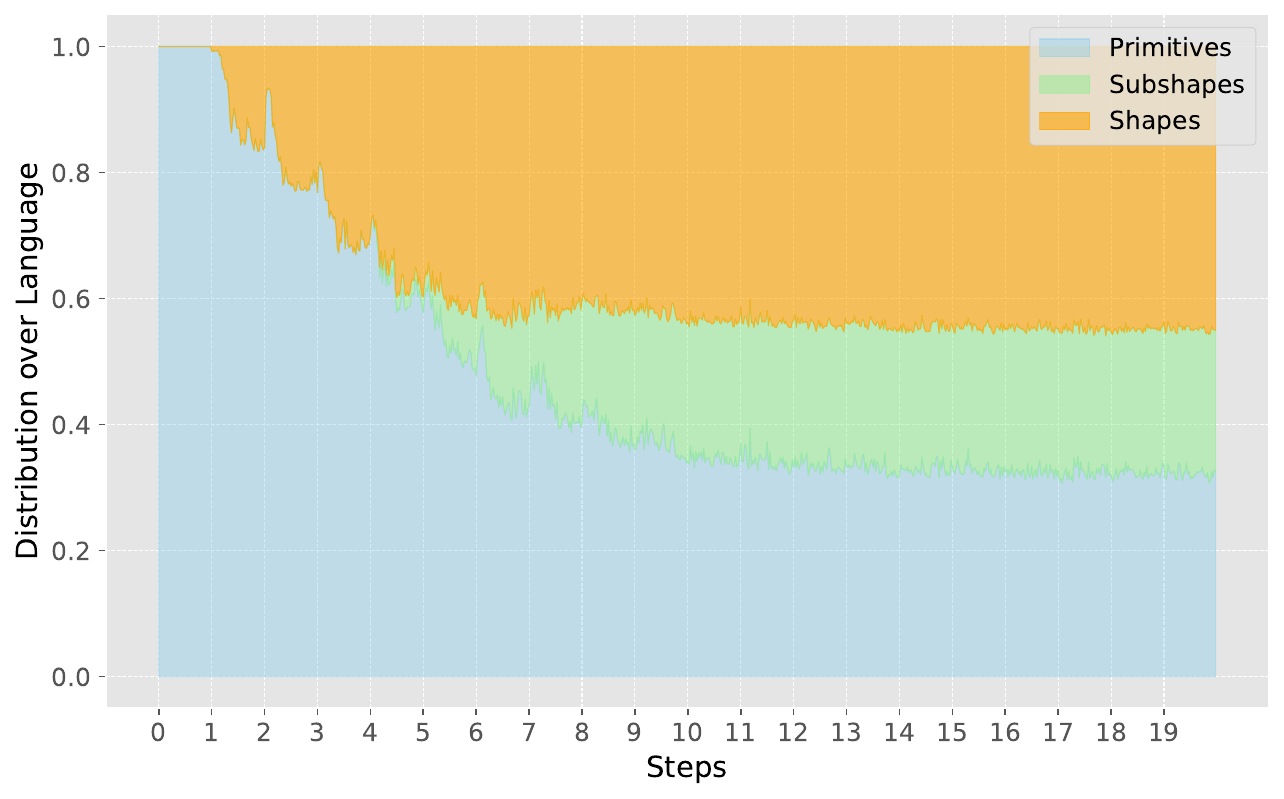}
    \caption{Relative proportion of actions as the language changes over training. Primitives refer to the initial actions, shapes refer to one of the 31 discrete shapes appearing in goal-scenes, and sub-shapes refer to anything else. }
    \label{fig:language_distribution}
\end{figure}

\begin{figure}[htb]
    \centering
    \includegraphics[width=\linewidth]{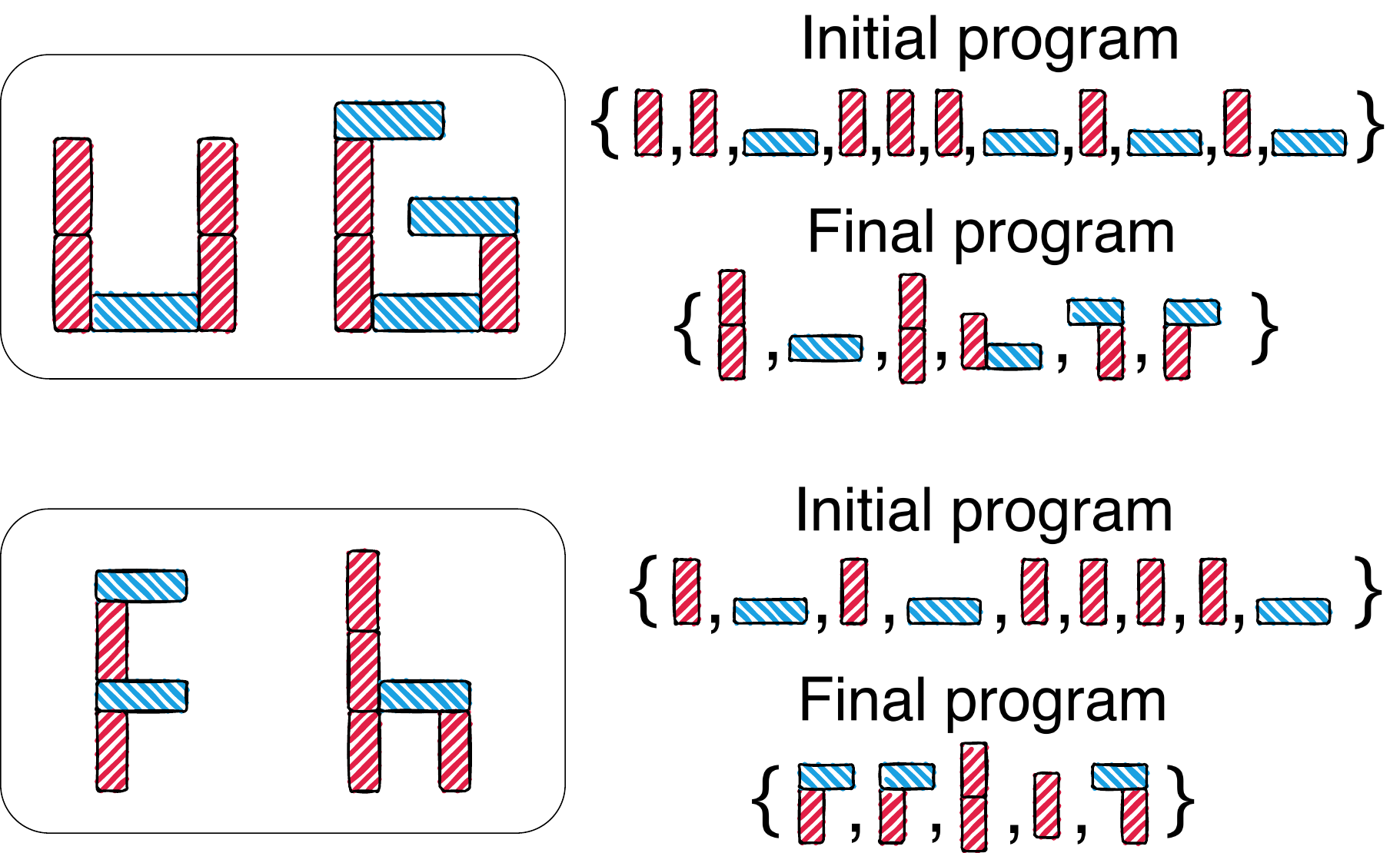}
    \caption{Two goal-scenes with representations of their initial and final programs chosen by PACE.}
    \label{fig:goal_scenes}
\end{figure}

In Figure \ref{fig:language_distribution}, we show how the composition of our language changes throughout the repeated interactions. The relative proportion of primitives is reduced in favour of abstractions referring to either one of the 31 discrete shapes in the dataset (roughly 40\%), or a sub-shape (roughly 20\%). \textcolor{black}{We show how this change in distribution impacts the composition of programs in Figure \ref{fig:goal_scenes} with two goal-scenes examples. We see that the introduction of new abstractions enables for programs to be rewritten much more compactly.}

%the  we find that it reduces the relative proportion of primitives, and favours more complex shapes which enable shorter program lengths. 

\section{Related Literature}

\citet{McCarthyBuilding21} propose a Bayesian model of procedural abstractions that combines library learning \citep{ellis2021dreamcoder} with social reasoning-based communication \citep{goodman2016pragmatic}. While effective in capturing human-like trends, this model assumes a predefined mapping from instructions to builder's actions, essentially defining a priori the meaning of messages that might not have been introduced yet. They also explicitly constrain language size to prevent adding further abstractions after a certain number have been introduced. \citet{jergeus2022towards} take a Deep Reinforcement Learning approach but oversimplify interactions by assuming the builder immediately understands abstractions, collapsing the setup into a single-agent framework and removing pressures for efficient communication. By using EC within PACE, our approach enables both agents to learn interactively, introducing abstractions organically and letting communicative pressures naturally limit language size.

Emergent Communication has been widely used to study various human language phenomena, such as compositionality \citep{mordatch2018emergence, Chaabouni2020, Ren2020Compositional}, the impact of populations \citep{kim2021emergent, chaabouni2022emergent, rita2022on}, and naming conventions in semantic categories \citep{carlsson2024cultural}. The architect-builder game presents a very challenging setting, being a repeated game with multiple interactions per round and sparse rewards. By incorporating symbolic methods with emergent communication, we provide a more interpretable, easy-to-use approach for modelling the introduction and use of procedural abstractions into conversational AI dyads.%In preliminary experiments, we found that direct application of EC did not produce conversational characteristics similar to those in human studies, and the outcomes lacked interpretability. By incorporating symbolic methods we provide a more interpretable, easy-to-use approach for modelling procedural abstractions.

\section{Conclusion}

\begin{comment}
Our model provides a model of human communication.
It provides a more natural model that automatically converges to a stable language. Reduces dependency on w. Connections to efficient communication.
Limitations. This is an instructional view of abstraction, but doesn't capture how this may extend to natural language. Will explain this, appreciate it is unclear.
Future Work.

The objective of this paper was to explore whether communication could shape the learning of abstractions as has been hypothesised by \cite{Ho}. Our results provide support for this hypothesis. They demonstrate how communication influence the way an agent constructs programs to solve collaborative tasks. In line with results from \cite{}

\end{comment}

In collaborative task-orientated communication humans tend towards more concise utterances by introducing procedural abstractions. In this work we propose a novel neuro-symbolic algorithm called PACE which displays similar tendencies. Our work serves as a bridge between procedural abstraction learning and efficient communication \citep{Kemp2012, Gibson2016, Zaslavsky2019a, Gibson2019, denic2024recursive, carlsson2024cultural}. We demonstrate that more optimal languages are easier to learn reinforcing ideas relating learnability with efficient communication. 

In future work, we intend to extend our analysis to consider other collaborative domains and deepen the comparison to human behaviour. We also intend to explore how Large Language Models 
understand and reason about these conversational dynamics. Significant research effort is being invested into exploring the role of natural language as a mechanism for humans to provide instructions to intelligent agents \citep{brohan2023can, shi2024yellrobotimprovingonthefly}. Providing these systems with the capability of handling novel procedural abstractions can facilitate improved cooperation between humans and agents. We believe that PACE represents a step towards equipping intelligent agents with flexible and extendable languages. % 
\begin{comment}
    
So far, we have not applied ACE to domains beyond the Architect-Builder game, and an extension towards novel environments like ARC \cite{chollet2019measure} might mean that the assumption that a set of initial programs exists for each goal-state may not be realisable, further complicating the communicative task. Furthermore, our work does not yet analyse ACE's languages from an information-theoretic perspective, e.g. as a trade-off between average program length and vocabulary size, situating them in relation to a \emph{Pareto frontier}.
\end{comment}

\section{Acknowledgments}

This work was supported by funding from from the Wal-
lenberg AI, Autonomous Systems and Software Program
(WASP) funded by the Knut and Alice Wallenberg Foun-
dation and the Swedish Research Council. The computa-
tions in this work were enabled by resources provided by
the Swedish National Infrastructure for Computing (SNIC).
We also extend our thanks to Emil Carlsson and Vikas Garg for insightful
discussions at various points throughout the project and to Sandro Stucki for valuable comments on a previous version of the manuscript that lead to this final one.

\setlength{\bibleftmargin}{.125in}
\setlength{\bibindent}{-\bibleftmargin}

\bibliography{refs}

\appendix
\onecolumn
\section*{Appendix}

\subsection{Algorithms composing PACE}
\label{app:algos}

Algorithm \ref{alg:em_abstract} shows how the components of PACE integrate. The algorithm alternates between the \emph{Communication Phase} and the \emph{Abstraction Phase} for $s$ steps, where $s$ is determined experimentally\footnote{We found that 20 steps were a good choice for our setting, after which the language stabilises and no more abstractions are introduced in the emergent language.}. 

The \emph{Communication Phase} is repeated for $e$ epochs, determined experimentally. At each iteration, programs are sampled from $\pi_{arch}$ (the table of programs for scenes) by the bandit's $\epsilon$-greedy strategy and are used to form a dataset of grid-state transitions. This dataset forms the basis of our signalling game and is used to train $\pi_{comm}$, $\pi_{bldr}$ and to update the $Q$-values of actions. Here, we expect a language to emerge which is subject to pressures for efficient communication. % mention bandit here 

In the \emph{Abstraction Phase} sub-sequences in the programs are identified. The sub-sequence that maximises Equation \ref{eq:bayes_update} is selected. This abstraction, $a_{new}$, is introduced into $\cal A$. $a_{new}$ may allow existing programs to be rewritten: in this case, the resulting shorter programs are introduced into $\pi_{arch}$ for the respective goal-scene. Finally, $Q(a_{new})$ is initialised. 

\begin{algorithm}[htbp]
\caption{Procedural Abstractions for Communicating Efficiently (PACE)}
\label{alg:em_abstract}
\begin{algorithmic}[1]
\REQUIRE initial $\pi_{arch}$, initial $\cal A$, $q_{init}$, $\alpha$, $\epsilon$, $e$, $s$
\ENSURE trained $\pi_{comm}$, $\pi_{bldr}$, $Q$, updated $\pi_{arch}$, $\cal A$
\STATE Initialise neural networks $\pi_{comm}$ and $\pi_{bldr}$
\STATE Initialise action Q-values $Q(a)$ to $q_{init}$, $\forall a \in \mathcal{A}$ 
\FOR{$s$ steps}
 \STATE \textit{// Communication Phase}
    \FOR{$e$ epochs}
        \STATE For each goal choose a program in $\pi_{arch}$ via bandit
        \STATE Create a training dataset $D$ of triples $(x, a, x')$ of transitions based on chosen programs
        \STATE Play signalling games with $D$ to train $\pi_{comm}$, $\pi_{bldr}$ and $Q$-values
        %\STATE Play signalling game with $\pi_{comm}$ \& $\pi_{bldr}$
        %\STATE Update $\pi_{comm}$, $\pi_{bldr}$ and $Q$ values on game
    \ENDFOR
    \STATE Optional: evaluate on hold-out test dataset
\STATE \textit{// Abstraction Phase}
%\STATE Find the most common action subsequences 
%\STATE Create a new abstraction $a_{new}$ using (5)
%\STATE Compress existing programs using A if possible
%\STATE Add new compressed programs to the set of programs associated with each goal
\STATE Find action sub-sequences in programs
\STATE Choose new abstraction $a_{new}$ that maximises Eq. \ref{eq:bayes_update}
\STATE Compress existing programs using $a_{new}$, if possible
\STATE For each goal, add new programs to $\pi_{arch}$
%\STATE Add new programs to $\pi_{arch}$
\STATE Add $a_{new}$ to $\cal A$ and initialise $Q(a_{new})$ to $q_{init}$
\ENDFOR
\end{algorithmic}
\end{algorithm}   

We also provide more detailed algorithms for the focal points of PACE's methodology. Algorithms \ref{alg:dataset_gen} and \ref{alg:training} refer to the \textit{Communication Phase}, which we break down further into two parts: the dataset generation for the signalling game and the communication round done via EC. Algorithm \ref{alg:find_and_imp_abs} instead refers to the \textit{Abstraction Phase}.   

\begin{algorithm}
\caption{Dataset generation via bandit}
\label{alg:dataset_gen}
\begin{algorithmic}[1]
\REQUIRE $\pi_{arch}$, $Q$, bandit strategy
\ENSURE $D$ \textit{\# A dataset of signalling games}
\STATE $D = \{\}$ 
\FOR{$g \in G$}
    \STATE Get set of programs for $g$ from $\pi_{arch}$
    \STATE Compute Q-values for programs
    \STATE Pick program, $p$, via bandit strategy
    \STATE break $p$ in a set of triples $(x, a, x')$ of transitions 
    \STATE $D \leftarrow D \bigcup \{ (x_i, a_i, x'_i)\}_{i \in |p|}$
    
\ENDFOR

\end{algorithmic}
\end{algorithm}   

\begin{algorithm}
\caption{A round of communication of PACE}
\label{alg:training}
\begin{algorithmic}[1]
\REQUIRE $D$, $\pi_{comm}$, $\pi_{bldr}$, $Q$
\ENSURE trained $\pi_{comm}$, $\pi_{bldr}$, $Q$
\STATE \textit{\# Play signalling game and update policies}
\WHILE{$D$ is not empty}
\STATE Sample batch $B$ from $D$ without replacement 
\STATE $loss = 0$ 
\FOR{$(x, a, x') \in B$} 
    \STATE $m = \pi_{comm}(a)$ %change a
    %\STATE $m_l = \pi_{pos}(l)$ 
    \STATE $\widehat{x'} = \pi_{bldr}(x, m)$
    \STATE $loss \leftarrow loss + \mathcal{L}(x', \widehat{x'})$
    \STATE Update $Q(a)$ for all $a$ in $B$   
\ENDFOR
\STATE Update $\pi_{comm}$ and $\pi_{bldr}$ via gradient descent
            \ENDWHILE
\end{algorithmic}
\end{algorithm}   

\begin{algorithm}
\caption{Abstraction phase}
\label{alg:find_and_imp_abs}
\begin{algorithmic}[1]
\REQUIRE $\pi_{arch}$, $\cal A$
\ENSURE updated $\cal A$ and $\pi_{arch}$
\STATE $Z = \{\}$ \textit{\# The set of preferred programs}
\FOR{$g \in G$}
    \STATE Get set of programs for $g$ from $\pi_{arch}$
    \STATE Compute Q-values for programs 
    \STATE Pick programs, $p$, via greedy bandit strategy
    \STATE $Z \leftarrow Z \bigcup \{p\}$
\ENDFOR
\STATE Find all sub-sequences in $Z$ (candidate abstractions)
\STATE Select best abstraction, $a_{new}$, via Equation \ref{eq:bayes_update}
\STATE $A \leftarrow A \bigcup \{a_{new}\}$ 
\FOR{$g \in G$}
    \FOR{$p \in \pi_{arch}(g)$}
        \IF{$a_{new}$ is applicable to $p$}
            \STATE Add $p$ re-written in terms of $a_{new}$ to $\pi_{arch}(g)$
        \ENDIF
    \ENDFOR
\ENDFOR

\end{algorithmic}
\end{algorithm}

\subsection{Q-value initialisation: Pessimism is best}
\label{app:q_initialisation}

\begin{figure}[h]
    \centering
    \includegraphics[width=0.8\linewidth]{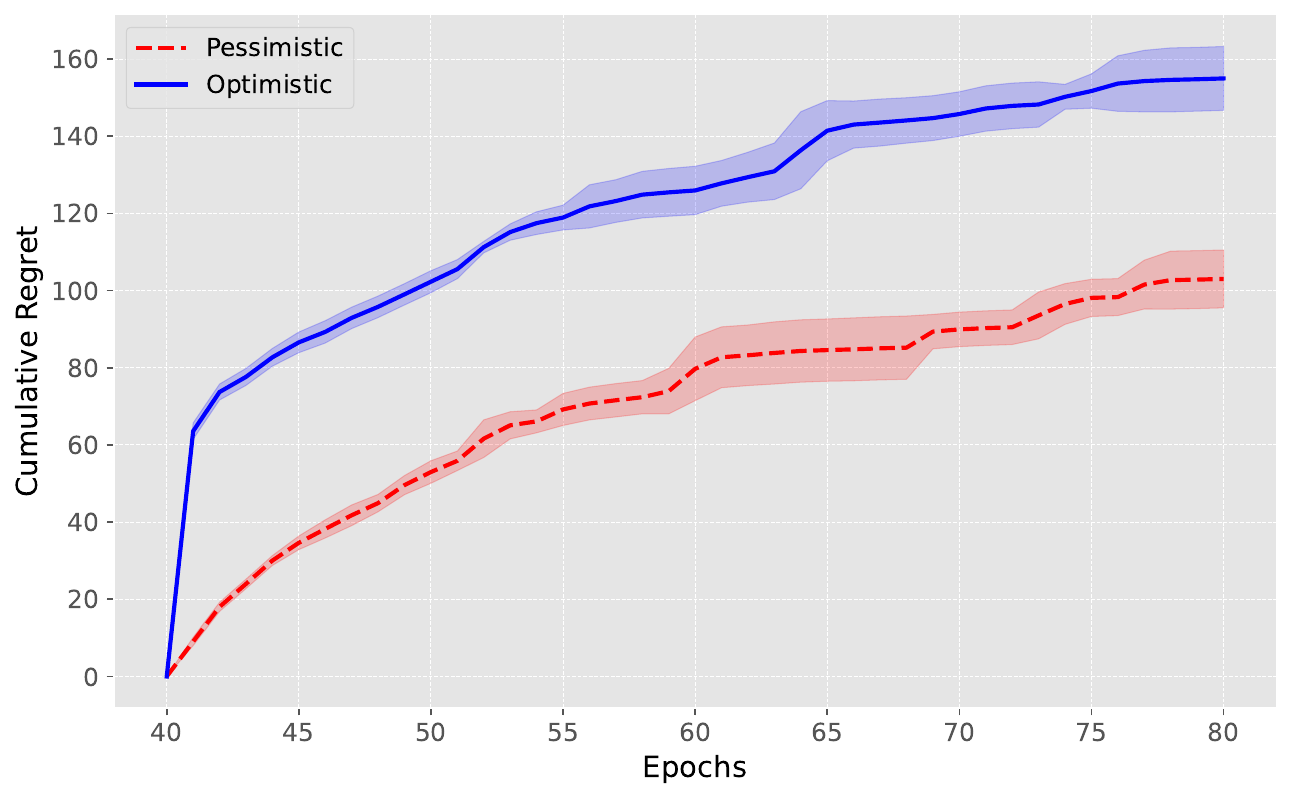}
    \caption{Cumulative regret for an optimistic and pessimistic initialisation of Q-values. Line indicates mean performance and shaded regions indicate the 95\% confidence interval for $16$ random seeds. Epoch interval is between step $1$ and step $2$.}
    \label{fig:opt_vs_pess}
\end{figure}

It is not obvious how the initialisation of the Q-values for new abstractions may impact regret. Compelling cases can be made for both optimistic (a more aggressive introduction) and pessimistic (a more gradual introduction) strategies. Empirically we find the pessimistic strategy works better in our case. Here, we establish its impact on learning of the common ``tower'' abstraction. 
We compare a pessimistic strategy ($q_{init}=0$) to an optimistic strategy ($q_{init}=1$) in terms of cumulative regret, defined as $Regret_T = \sum_{t=1}^T \left( r^* - r_{t} \right)$. In Figure \ref{fig:opt_vs_pess}, we see that a pessimistic initialisation of $0$ incurs significantly less cumulative regret than an optimistic value of $1$. This is attributable to the optimistic variant's tendency to force the abstraction into use before the builder has had the opportunity to become familiar with it. The pessimistic agent introduces the new abstraction more gradually while still providing the opportunity for learning of a common language.

\subsection{Examples of Abstractions Introduced}
\label{app:results}
\begin{figure}[htb]
    \centering
    \includegraphics[width=0.65\linewidth]{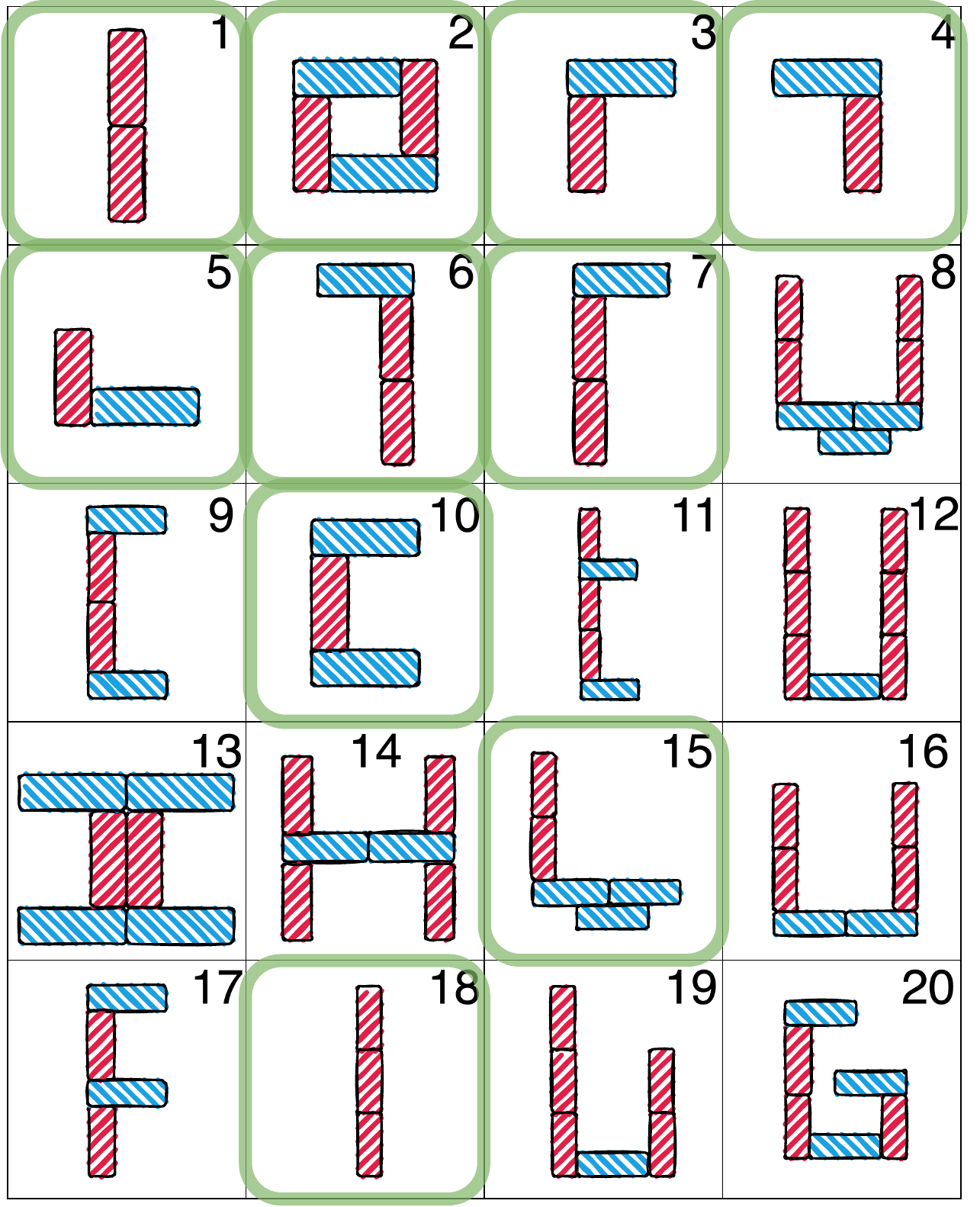}
    %\includesvg[width=0.65\linewidth]{images/abstractions_greenonly.svg}
    \caption{The abstractions introduced by PACE in an example run. We circle in green the abstractions that are adopted by the architect in its final language.}
    \label{fig:abstr_coloured}
\end{figure}

In Figure \ref{fig:abstr_coloured} we report the abstractions that PACE introduces in a single run. Abstractions circled in green are in the final language.

%the  we find that it reduces the relative proportion of primitives, and favours more complex shapes which enable shorter program lengths. 

\begin{figure}[p]
    \centering
    \includegraphics[width=0.65\linewidth]{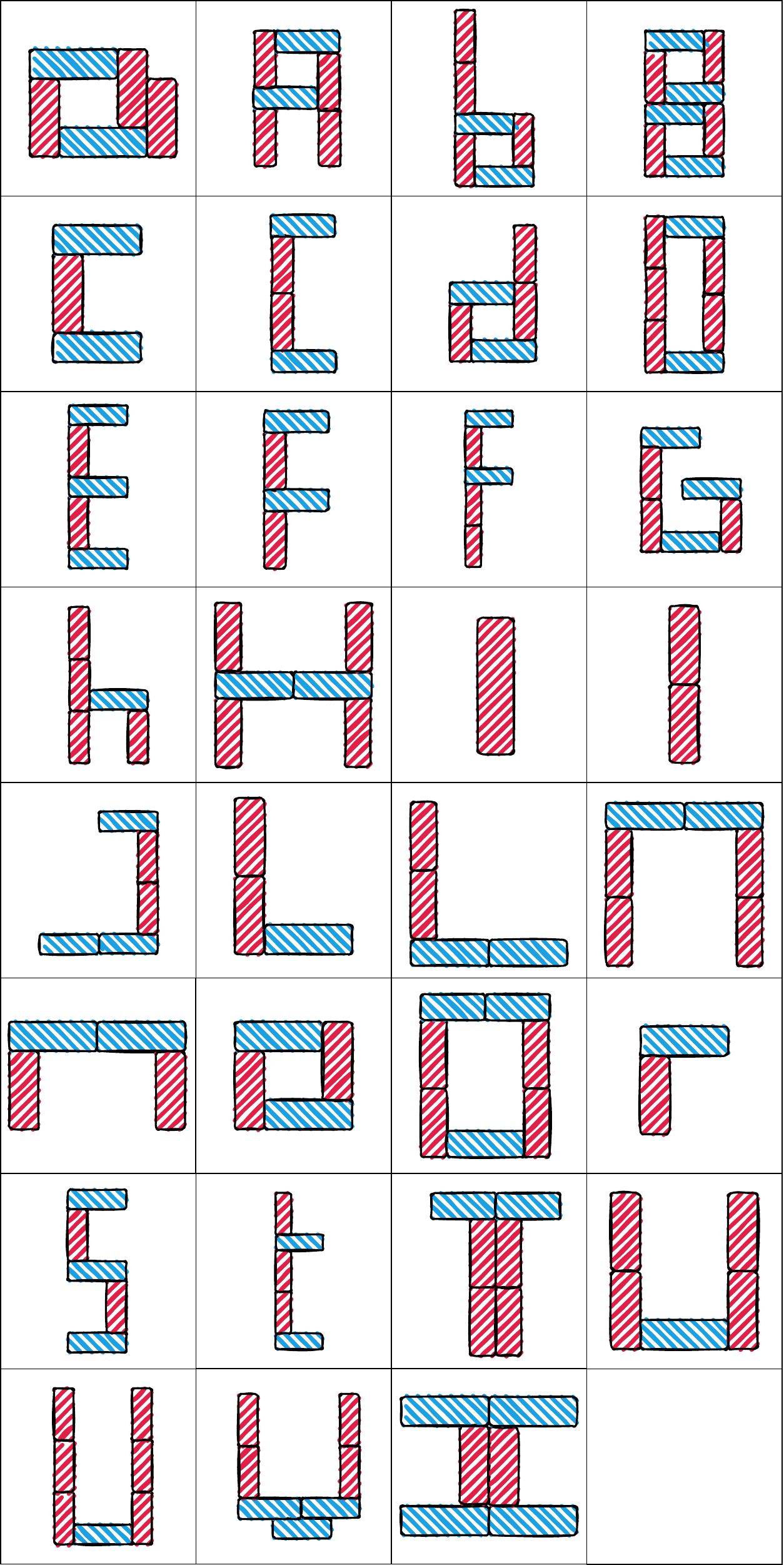}
      %\includesvg[width=0.65\linewidth]{images/dataset.svg}

    \caption{The $31$ shapes that in pair compose each goal-scene.}
    \label{fig:dataset_shapes}
\end{figure}

\subsection{Dataset Composition}
\label{app:dataset}

Each goal-scene in the dataset is composed of two shapes placed next to each other, as in Figure \ref{fig:arch_build_game}. These are sampled from 31 shapes we design and report in Figure \ref{fig:dataset_shapes}. Each shape is meant to resemble an uppercase or lowercase letter from the English alphabet. We note that there are multiple subshapes that recur several times within the 31 shapes, making our dataset a prime candidate for studying the introduction of abstraction.

\end{document}